\documentclass[journal]{IEEEtran}

\usepackage[pdftex]{graphicx}
\usepackage[caption=false]{subfig}

\usepackage{multicol}

\usepackage{amsmath,amssymb,bm}
\DeclareMathOperator*{\argmin}{arg\!\min} %

\usepackage{multirow}
\usepackage[hidelinks]{hyperref}

\usepackage[ruled,norelsize]{algorithm2e}
\makeatletter
\newcommand{\removelatexerror}{\let\@latex@error\@gobble}
\makeatother

\begin{document}
\title{RAPDARTS: Resource-Aware\\Progressive Differentiable Architecture Search}

\newcommand{\printfnsymbol}[1]{%
  \textsuperscript{\@fnsymbol{#1}}%
}

\author{
	\IEEEauthorblockN{Sam Green\IEEEauthorrefmark{3}\IEEEauthorrefmark{1}, Craig M. Vineyard\IEEEauthorrefmark{3}, Ryan Helinski\IEEEauthorrefmark{3}, \c{C}etin Kaya Ko\c{c}\IEEEauthorrefmark{1}}\\
	\IEEEauthorblockA{\IEEEauthorrefmark{3}Sandia National Laboratories, Albuquerque, New Mexico, USA
		\\\{sgreen, cmviney, rhelins\}@sandia.gov}\\
	\IEEEauthorblockA{\IEEEauthorrefmark{1}University of California Santa Barbara, Santa Barbara, California, USA
		\\cetinkoc@ucsb.edu}
}

\maketitle

\begin{abstract}
Early  neural network architectures were designed by so-called ``grad student descent''. Since then, the field of Neural Architecture Search (NAS) has developed with the goal of algorithmically designing architectures tailored for a dataset of interest. Recently, gradient-based NAS approaches have been created to rapidly perform the search. Gradient-based approaches impose more structure on the search, compared to alternative NAS methods, enabling faster search phase optimization. In the real-world, neural architecture performance is measured by more than just high accuracy. There is increasing need for efficient neural architectures, where resources such as model size or latency must also be considered. Gradient-based NAS is also suitable for such multi-objective optimization. In this work we extend a popular gradient-based NAS method to support one or more resource costs. We then perform in-depth analysis on the discovery of architectures satisfying single-resource constraints for classification of CIFAR-10.
\end{abstract}

\section{Introduction}
\IEEEPARstart{T}HE optimal design of a neural architecture depends on 1) the target dataset, 2) the set of \textit{primitive operations} (e.g. convolutional filters, skip-connections, nonlinearity functions, pooling), 3) how the primitive operations are composed into a neural architecture and optimized, and 4) resource constraints like hardware cost, minimum accuracy, or maximum latency. In this paper, we assume the target dataset has been provided, and we provide guidelines and analysis for searching for neural architectures under one or more hardware resource constraints.

Convolutional layers and fully-connected layers are parameter-heavy operations. Those, along with other lighter primitive operations, like pooling layers or batch normalization, may be composed into an endless variety of neural architectures. But what is the optimal neural architecture for a given dataset? There is no existing closed-form solution to that question. 

Historically, the highest performing neural architectures have been found by applying heuristics and a large amount of compute. Some well known examples of modern hand-crafted architectures include AlexNet \cite{krizhevsky2012imagenet}, VGG16 \cite{simonyan2014very}, ResNet \cite{he2016deep}, and the Inception series \cite{szegedy2015going, szegedy2016rethinking, szegedy2017inception}. None of these examples consider hardware, and they pursue classification performance at all cost. 

\textit{Neural Architecture Search} (NAS) methods automate strategies for discovery of high performing neural architectures. A reinforcement learning-based (RL) approach was the first post-AlexNet NAS method with state-of-the-art performance on CIFAR-10 \cite{zoph2016neural, cifar10}. The RL approach was quickly followed by a high performance Evolutionary Strategy (ES) based method \cite{real2017large}. While both the RL and ES methods discovered high performance architectures, their use came at the cost of thousands of GPU hours.

\begin{figure}[t]
\centering
\includegraphics[width=3.00in]{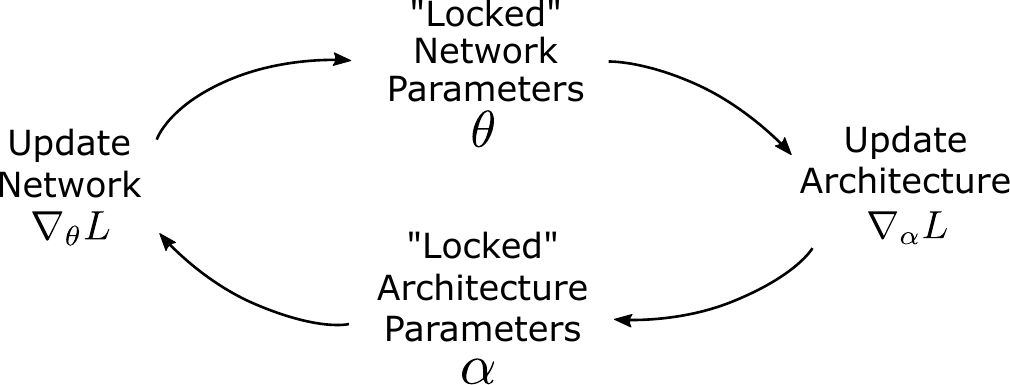}
\caption{Gradient-based Neural Architecture Search (GBNAS) methods maintain two sets of parameters. \textit{Neural network parameters} are represented by $\bm{\theta}$ and \textit{architecture parameters} are represented by $\bm{\alpha}$. GBNAS algorithms leverage differentiable functions, parameterized by architecture parameters, to design deep neural networks, which are parameterized by network parameters. First-order optimization alternates between ``locking'' one set of parameters and updating the other.}
\label{fig:cycle}
\end{figure}

\textit{Gradient-based NAS} (GBNAS) methods have the benefit of being directly optimized through gradient descent and consequently complete the search faster than other NAS methods. The basic idea of GBNAS is given in Figure \ref{fig:cycle}. The search process alternates between temporarily fixing one set of parameters, i.e. assuming they are constants, and updating the other set of parameters. This approach has no convergence guarantees, but it works well in practice.

Because neural models are now widely deployed on systems like edge devices, in cars, and running in servers, available hardware resources also have an impact on what may be considered an ``optimal'' neural architecture design. Hardware resource constraints are often summarized as size, weight, and power (SWaP). Resource constraints could also include maximum latency, minimum throughput, or a manufacturing budget which will determine if a custom ASIC is an option, if a COTS device is sufficient, or if something semi-custom, like an FPGA, is an option. For example, during the design of Google's TPUv1, architects were given a budget of 7 ms per inference (including server communication time) for user-facing workloads \cite{jouppi2017datacenter}.

Recent efforts described below implement NAS strategies incorporating hardware resource constraints into the search. GBNAS methods capture hardware resource constraints within a differentiable loss function. This approach enables the architecture search to yield network architectures biased toward satisfying resource constraints.

In this work we have modified P-DARTS \cite{chen2019progressive}, which in-turn is based on another popular gradient-based NAS algorithm, DARTS \cite{liu2018darts}, to support resource costs. We use our modified GBNAS algorithm to search for many neural architectures under various resource consumption penalties. We then use our results and observations to answer the following questions:
\begin{itemize}
\item What is the computational cost of searching for satisficing architectures?
\item What heuristics can be used to guide the search and training process to reduce compute time?
\item How reproducible are search results under random initial conditions?
\end{itemize}
\section{Related Work}

\begin{figure}[!t]
\centering
\includegraphics[width=2.75in]{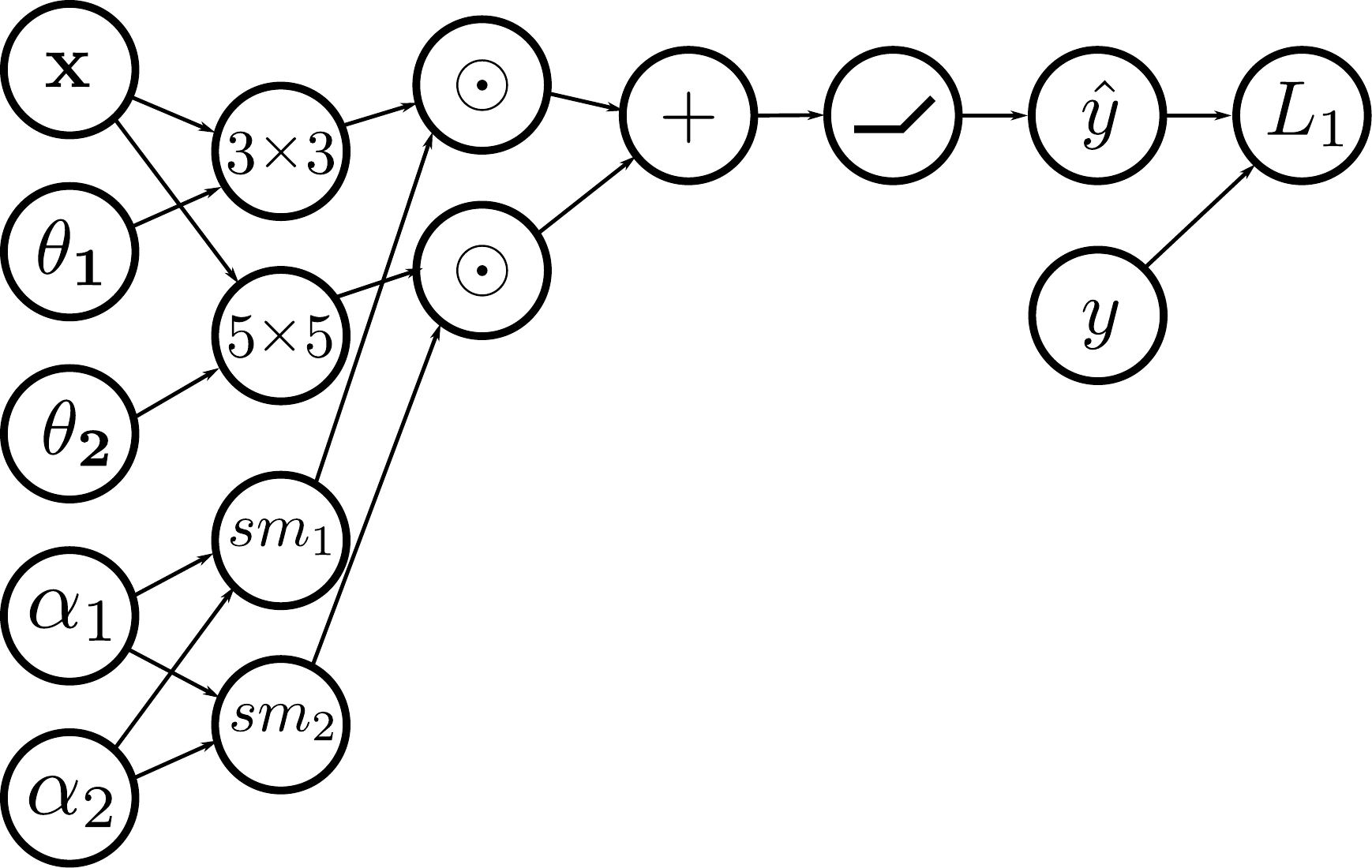}
\caption{The function of DARTS architecture parameters is to scale the output of primitive operations. In this illustration the primitive operations include $3 \times 3$ and $5 \times 5$ convolutional filters parameterized by tensors $\bm{\theta}_1$ and $\bm{\theta}_2$ respectively. The output feature maps of the primitive operations are element-wise scaled $\odot$ by the softmax ($sm$) of architecture parameters $\alpha_1$ and $\alpha_2$. The scaled output feature maps are then added, thereby creating a \textit{mixed operation}. This notional illustration shows a network with only two primitive operations, followed by a nonlinearity, producing an output prediction $\hat{y}$. In practice, there may be many mixed operations, each containing many primitive operations, forming a deep network.}
\label{fig:gbnas}
\end{figure}

The first competitive NAS approach applied to modern image classification tasks was based on reinforcement learning (RL) \cite{zoph2016neural}. In this work, an LSTM-based RL agent was trained to output primitive operations which were then chained together into a directed acyclic graph. After training and evaluating the graph, the agent was then encouraged or discouraged, via a positive or negative reward derived from classification accuracy, to generate similar graphs in the future or to explore and make new graphs. 

The reinforcement learning NAS approach worked well and was able to achieve high accuracy, but at unheard of computational expense. It required 3,150 GPU-days to discover one of their published architectures.

Related approaches to sampling neural architectures include Markov chain Monte Carlo methods \cite{smithson2016neural}, evolutionary strategies \cite{elsken2018efficient}, and genetic algorithms \cite{lu2019nsga}. Similar to RL approaches, all of these optimization methods generate populations of neural architectures. The populations are then trained and a fitness value is derived from the classifier's final test performance. The fitness value is used to encourage or discourage the design of the next population of architectures.

Reinforcement learning, Markov chain Monte Carlo methods, evolutionary strategies, and genetic algorithms discover high-performance architectures, but they are incredibly expensive. These methods often require 100$\times$ to 1000$\times$ more compute than gradient-based methods \cite{wistuba2019survey}.

Gradient-based neural architecture search has recently become popular because of its efficiency \cite{liu2018darts, xie2018snas, cai2018proxylessnas, chen2019progressive}. GBNAS methods maintain two sets of parameters: \textit{network parameters} $\bm{\theta}$ and architecture parameter $\bm{\alpha}$. Previous GBNAS methods have introduced various methods to optimize and use the two parameter sets. In the simplest case, optimization is achieved by optimizing one set of parameters and then the other. This first-order optimization approach is illustrated in Figure \ref{fig:cycle}.

\begin{figure}[!t]
\centering
\includegraphics[width=2.35in]{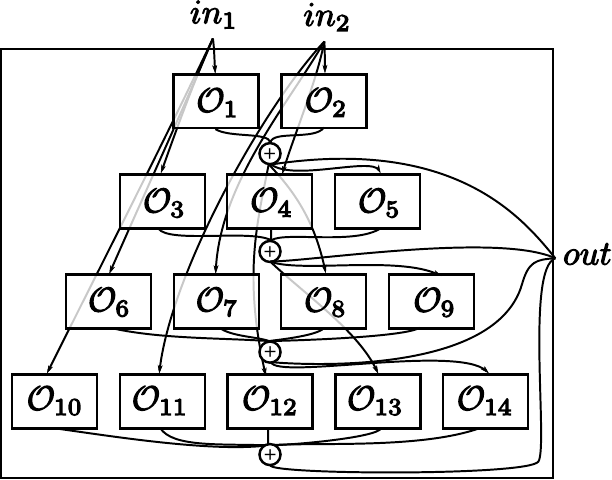}
\caption{The DARTS cell architecture has 14 mixed operations (represented by $\mathcal{O}_i$) distributed among four steps with skip-connections between each step. At each step, the outputs of the mixed operations are element-wise added. The sum is then passed as an input to a mixed operation in the next step. All element-wise sums are concatenated as the cell output and fed forward to the next cell in the network.}
\label{fig:cell}
\end{figure}

Differentiable Architecture Search (DARTS) is a GBNAS technique that uses \textit{mixed operations} to compute multiple primitive operations in parallel, followed by element-wise summation \cite{liu2018darts}. The mixed operations are scaled by architecture parameters prior to summation. For example, as illustrated in Figure \ref{fig:gbnas}, a $3\times3$ convolutional filter and a $5\times5$ convolutional filter can be designed such that both receive the same input feature map and both generate additively conformable output feature maps.

Extending this technique, DARTS composes 14 mixed operations into a \textit{cell}. Eight cells are then chained to create the network. Each cell has the same connectivity and architecture parameters ($\bm{\alpha}$) for mixed operations, but the network parameters ($\bm{\theta}$) are learned independently in each primitive operation and in each cell. An illustration of the DARTS cell connectivity is given in Figure \ref{fig:cell}.

DARTS has a limitation which requires the entire neural network (i.e. all cells and all mixed operations) to fit in GPU memory. This limits the depth of the neural network as well as the batch size during training. Progressive Differentiable Architecture Search (P-DARTS) mitigates the memory limitation of DARTS by 1) gradual growth in the depth of the neural network, and simultaneously 2) gradual reduction in number of primitive operations per mixed operation, thus reducing model size \cite{chen2019progressive}.

ProxylessNAS also extended DARTS \cite{cai2018proxylessnas}. ProxylessNAS treats the architecture parameters of each mixed operation as a probability distribution. ProxylessNAS stores a large overparameterized network in system memory, because the network is too large to fit on a GPU. During evaluation, a subnetwork is sampled and transferred to the GPU for evaluation. Gradients are calculated and used to update the shared-weights of the overparameterized network. 

Addressing the need to search for architectures which not only strive for high accuracy, but also meet additional performance constraints, hardware-aware NAS techniques have been pursued. ProxylessNAS is particularly relevant for hardware-aware GBNAS, because it formalizes the approach to incorporating resource costs during the search. In the context of classification, ProxylessNAS creates a loss function that incorporates both a cross-entropy loss for the classification accuracy as well as a resource loss for latency. 

In this work we augment P-DARTS with a ProxylessNAS-style resource loss and analyze its impact on architectures discovered during the search phase.

\section{Method}
\subsection{Resource-Aware Differentiable Neural Architecture Search}
When training a convolutional neural network for classification, the goal is to obtain a model that best predicts labels from observations drawn from an underlying distribution of interest. Fitting a neural model to an underlying distribution is achieved by finding optimal network parameters $\bm{\bm{\theta}^*}$ that minimize expected prediction error on an available dataset:
\begin{equation}
\bm{\bm{\theta}}^{*} = \argmin_{\bm{\theta}} \big[ J(\bm{\bm{\theta}}) = \mathbb{E}_{(\bm{x},y) \sim \hat{p}_{data}} L(f(\bm{x}; \bm{\theta}), y) \big],
\label{eq:simpleobjective}
\end{equation}
where $J$ is the objective function, $\bm{x}$ are dataset observations, $y$ are dataset labels, $\hat{p}_{data}$ is the empirical distribution, $L$ is a prediction error loss function, and $f$ is the neural network parameterized by $\bm{\theta}$.

Gradient-based NAS methods introduce another set of \textit{architecture parameters} $\bm{\alpha}$, producing:
\begin{equation}
g(\bm{x}; \bm{\theta}, \bm{\alpha}).
\label{eq:dual}
\end{equation}
We refer to $g$ as a directed acyclic graph, or simply \textit{graph}, to highlight that it is composed of a neural network whose control flow is modified by other non-network architecture parameters. Note the distinction between $f$ used in Equation \ref{eq:simpleobjective}, which is only parameterized by network parameters, and $g$ used in Equation $\ref{eq:dual}$, which is parameterized by both network and architecture parameters.

Architecture parameters, like network parameters, are scalar-valued tensors. Architecture parameters are used to control either the weight of primitive operations, as in \cite{liu2018darts, chen2019progressive}, or the probability primitive operations will take place, as in \cite{guo2019single, cai2018proxylessnas}. In both cases, the scalar values are at least interpreted as one or more probability distributions through processing by the softmax function. In our case, the probability distribution is then used for evaluation of a mixed operation.

A mixed operation is illustrated in Figure \ref{fig:gbnas}, and it is formalized as:
\begin{equation}
\mathcal{O}(\bm{x}) = \mathbb{E}\big[ o(\bm{x}) \big] \approx \sum_{i=1}^{N} \frac{\text{exp}(\alpha_i)}{\sum_j \text{exp}(\alpha_j)}o_i(\bm{x}) = \sum_{i=1}^{N} p_i o_i(\bm{x}),
\label{eq:expected}
\end{equation} 
where $o_i(\bm{x})$ is a primitive operation, and $\mathcal{O}(\bm{x})$ is equivalent to the expected value of the primitive operations. This formalism extends the mixed operation to the inclusion of $N$ primitive operations that are evaluated in parallel and designed such that their outputs are additively conformable. In practice many mixed operations are used, with unique subsets of $\bm{\alpha}$ and $\bm{\theta}$ used for the calculation of each expected value, but we show only a single mixed operation here for clarity.

The inclusion of architecture parameters implies there are now two objective functions to be optimized:
\begin{align}
J(\bm{\bm{\theta}}) &= \mathbb{E}_{(\bm{x},y) \sim \hat{p}_{data}} L_1(g(\bm{x}, \bm{\alpha}; \bm{\theta}), y), \nonumber \\
J(\bm{\bm{\alpha}}) &= \mathbb{E}_{(\bm{x},y) \sim \hat{p}_{data}} L_1(g(\bm{x}, \bm{\theta}; \bm{\alpha}), y).
\label{eq:two_obj}
\end{align}

The graph evaluations in Equation \ref{eq:two_obj} are now denoted $g(\bm{x}, \bm{\alpha}; \bm{\theta})$ and $g(\bm{x}, \bm{\theta}; \bm{\alpha})$. This notation highlights that in the case of $J(\bm{\theta})$ the graph is evaluated at input and architecture parameter constants $(\bm{x}, \bm{\alpha})$ and optimized using network parameters $\bm{\theta}$. In the second case of $J(\bm{\alpha})$ the graph is evaluated at input and network parameter constants $(\bm{x}, \bm{\theta})$ and optimized using architecture parameters $\bm{\alpha}$. Therefore the following bilevel optimization must be solved:
\begin{align}
\bm{\bm{\theta}}^{*} = \argmin_{\bm{\theta}} \big[ J(\bm{\bm{\theta}}) = \mathbb{E}_{(\bm{x},y) \sim \hat{p}_{data}} L_1(g(\bm{x}, \bm{\alpha}^*; \bm{\theta}), y) \big], \nonumber \\
\bm{\bm{\alpha}}^{*} = \argmin_{\bm{\alpha}} \big[ J(\bm{\bm{\alpha}}) = \mathbb{E}_{(\bm{x},y) \sim \hat{p}_{data}} L_1(g(\bm{x}, \bm{\theta}^*; \bm{\alpha}), y) \big].
\label{eq:bilevel}
\end{align}
When using first-order differentiable methods, this bilevel optimization is solved by alternatingly ``locking'' one set of parameters and updating the other with gradient descent. Second-order optimization methods, which involve calculation of the Hessian, are also possible and slightly better in terms of accuracy, but this comes at significant computational cost. However, it is possible to approximate the second-order optimization with reduced computational cost \cite{liu2018darts}.

\begin{figure}[!t]
\centering
\includegraphics[width=2.75in]{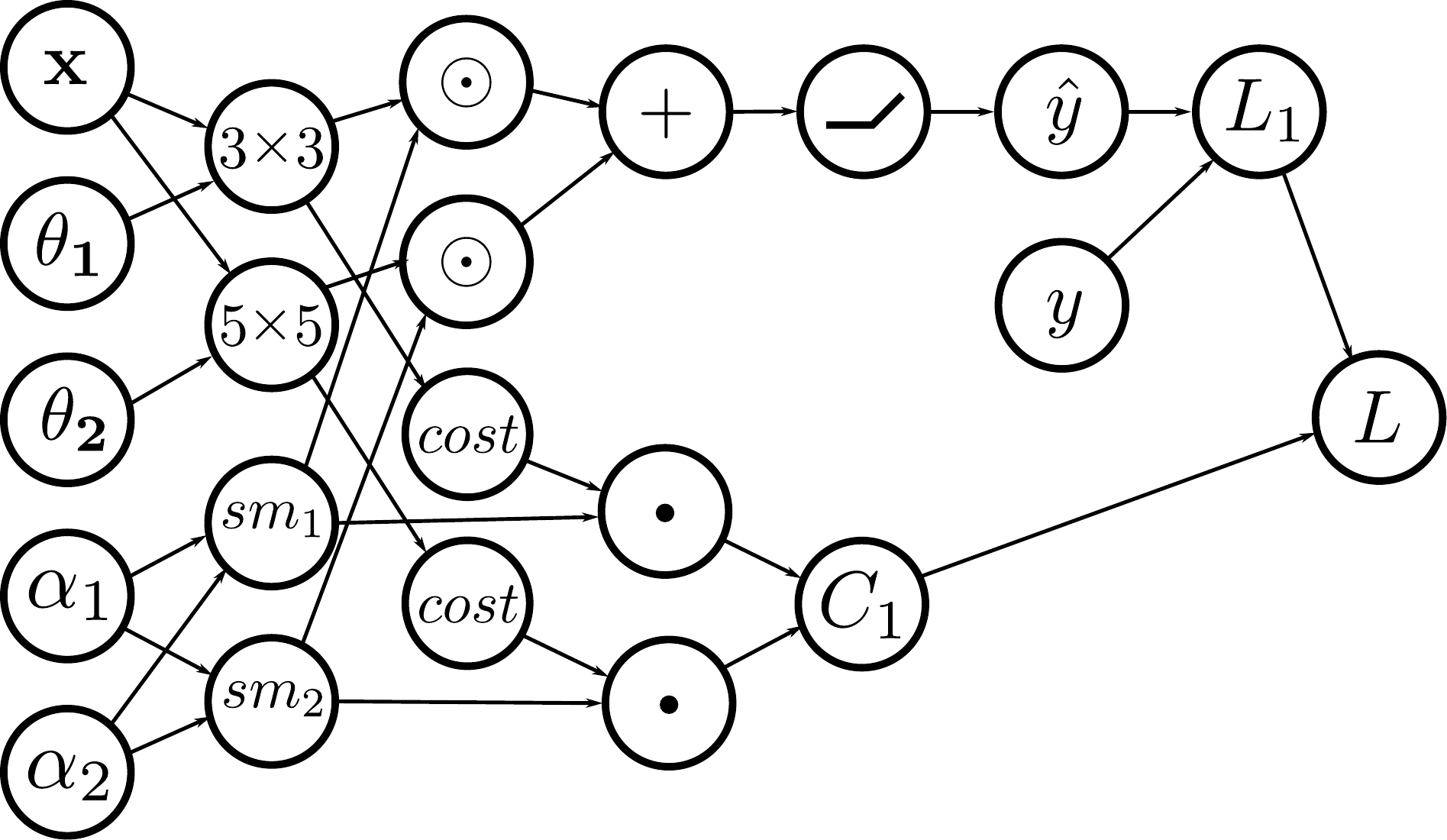}
\caption{P-DARTS may be extended with the calculation of an expected resource cost ($C_1$) for each mixed operation. When the gradient of the expected resource cost is calculated, the more expensive primitive operations are penalized more heavily than the less expensive operations, but the penalty is balanced by how much the primitive operation contributes to classification accuracy.}
\label{fig:RAPDARTS}
\end{figure}

Our method extends P-DARTS to discover neural architectures biased toward the satisfaction of resource constraints. We do this by including one or more ``expected resource cost'' loss terms. As mentioned previously, each of the primitive operations in a mixed operation is associated with a unique architecture parameter. P-DARTS uses 14 mixed operations in the search phase of cell architecture discovery, and there are eight primitive operations per mixed operation, so there are $14 \times 8 = 112$ architecture parameters total. 

The expected value of a single mixed operation was given in Equation \ref{eq:expected}. We temporarily make index values of the mixed operation explicit here for clarity:
\begin{equation}
\mathcal{O}_k(\bm{x}_k) = \sum_{i=1}^{8} p_{k,i} \cdot o_{k,i}(\bm{x}_k),
\label{eq:expectedk}
\end{equation}
where $k$ is the mixed operation index. Note here that the probability distributions, $p_{k,i}$, are now tied to a particular mixed operation. This calculation is equivalent to the addition node in Figure \ref{fig:gbnas}.

As introduced in ProxylessNAS, the probabilities used in the mixed operation calculation are also conducive to calculation of the expected value of various resource costs. For example, if there is a cost function that takes as input the description of each primitive operation (including the input feature map dimension information) and outputs a resource cost, it may be used for the calculation of an expected resource cost of the mixed operation:
\begin{equation}
\mathbb{E}\big[ cost(\mathcal{O}_k(\bm{x}_k)) \big] \approx \sum_{i=1}^{8} p_{k,i} \cdot cost(o_{k,i}(\bm{x}_k)).
\label{eq:expectedr}
\end{equation}

The cost function may be an analytical function, e.g. number of bytes required by the model, or the cost function could be based on a simulation or a surrogate model trained from data collected from a physical device. 

The expected cost of the mixed operation is differentiable with respect to the mixed operation's architecture parameters. Accordingly, the partial derivative of the expected resource cost with respect to architecture parameter $\alpha_i$ is given as:
\begin{align}
\frac{\partial \mathbb{E}\big[ cost(\mathcal{O}(\bm{x})) \big]}{\partial \alpha_i} &\approx \frac{\partial \big[p_1c_1 + p_2c_2 + \cdots + p_8c_8 \big]}{\alpha_i}, \nonumber \\
& = \sum_{l=1}^{8} \frac{\partial \Big[ \frac{\text{exp}(\alpha_l)}{\sum_j \text{exp}(\alpha_j)} \cdot c_l \Big]}{\partial \alpha_i}, \nonumber \\
& = \sum_{l=1}^{8} c_lp_l(\delta_{i,l} - p_i).
\label{eq:dexpectedr}
\end{align}
where we have abbreviated $cost(o_{i}(\bm{x}))$ as $c_i$, $\delta_{i,l} = 1$ if $i$ equals $l$ and $0$ otherwise, and we have dropped the mixed operation index $k$ for brevity.

We denote the sum of expected mixed operation costs as:
\begin{equation}
C_m = \sum_{k=1}^{14} \mathbb{E}\big[ cost_m(\mathcal{O}_k(\bm{x}_k)) \big],
\label{eq:sumexp}
\end{equation}
Note that unique $m$ correspond to unique resource costs, e.g. $C_1$ could be the sum of expected mixed operation parameter sizes, and $C_2$ could be the sum of expected mixed operation latencies.

We denote the sum of the classification and resource losses as:
\begin{equation}
L = L_1 + \sum_{m=1}^{M} \lambda_m C_m,
\label{eq:totalloss}
\end{equation}
where $M$ is the number of resource costs to satisfy, and $\lambda_m$ is the resource-cost hyperparameter and controls how important the resource cost $m$ is compared to accuracy as well as other resource costs.

\begin{figure}[!t]
\centering
\includegraphics[width=3.45in]{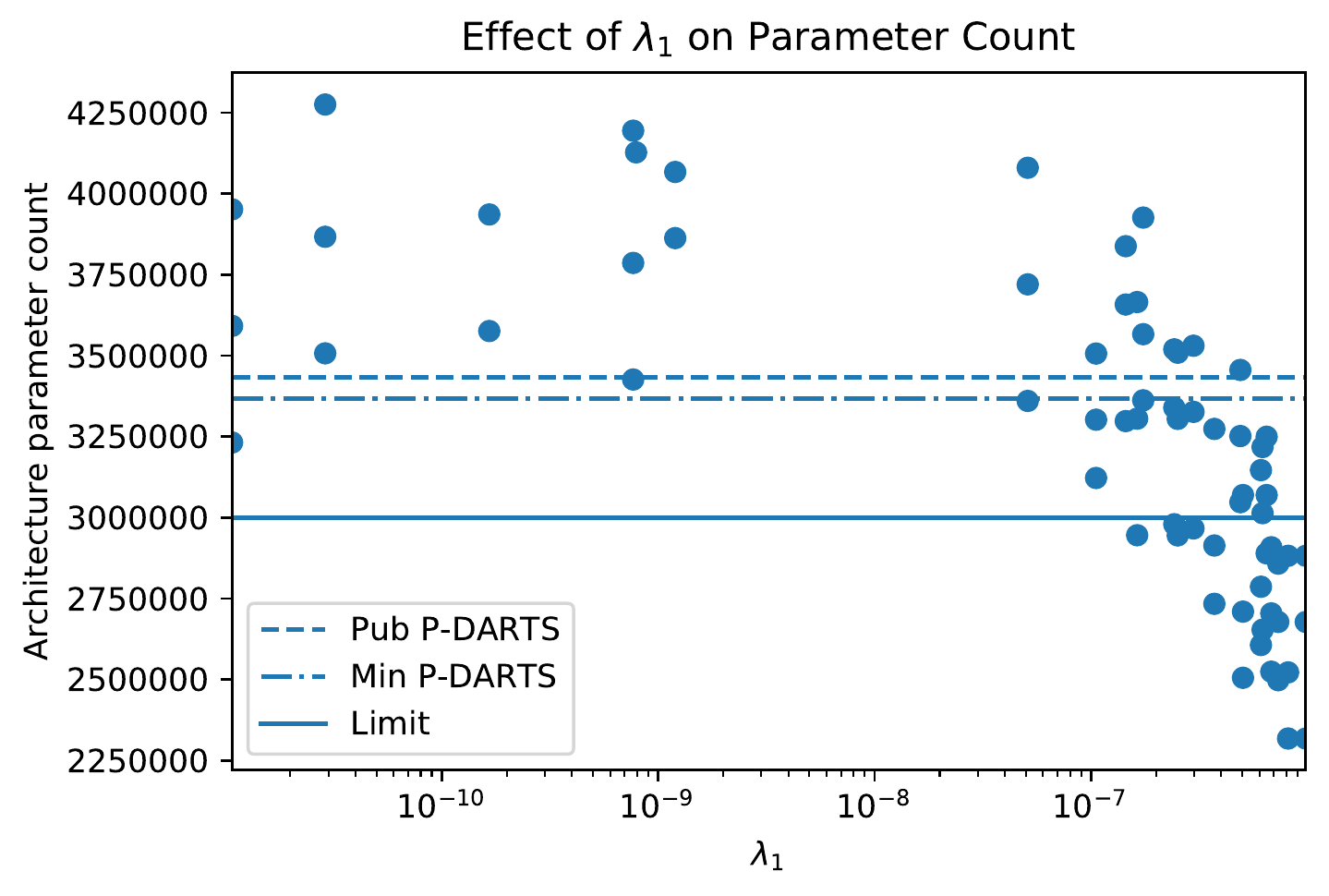}
\caption{Coarse-search for resource expected parameter count hyperparameter $\lambda_1$. As $\lambda_1$ grows beyond $10^{-7}$, RAPDARTS increasingly identifies architectures that require less than 3 M parameters. The publish P-DARTS architecture is marked with the dashed line. The minimum P-DARTS architecture found by us is marked with the dash-dot line. Our self-imposed budget is marked with the solid line.}
\label{fig:lam_bytes}
\end{figure}

\begin{figure}[!t]
\centering
\includegraphics[width=3.45in]{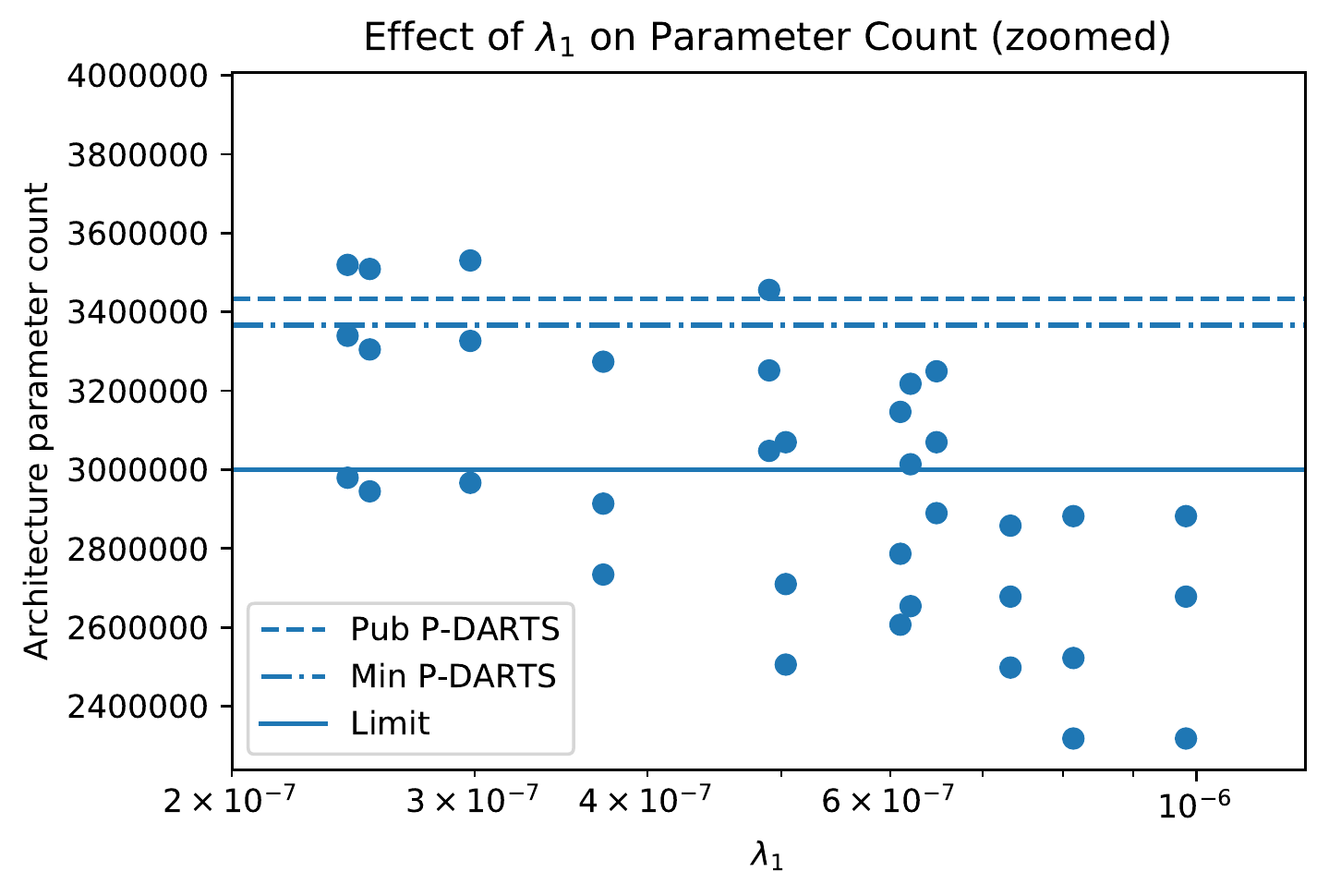}
\caption{Fine-search focused $2 \times 10^{-7} < \lambda_1 < 10^{-6}$. At around $\lambda_1 = 6 \times 10^{-6}$ architectures are frequently generated which meet the 3 M parameter constraint.}
\label{fig:lam_bytes_zoom}
\end{figure}

The bilevel optimization in Equation \ref{eq:bilevel} may now be slightly rewritten as:
\begin{align}
\bm{\bm{\theta}}^{*} = \argmin_{\bm{\theta}} \big[ J(\bm{\bm{\theta}}) = \mathbb{E}_{(\bm{x},y) \sim \hat{p}_{data}} L(g(\bm{x}, \bm{\alpha}^*; \bm{\theta}), y) \big], \nonumber \\
\bm{\bm{\alpha}}^{*} = \argmin_{\bm{\alpha}} \big[ J(\bm{\bm{\alpha}}) = \mathbb{E}_{(\bm{x},y) \sim \hat{p}_{data}} L(g(\bm{x}, \bm{\theta}^*; \bm{\alpha}), y) \big],
\label{eq:bilevel2}
\end{align}
where only $L_1$ has been replaced by $L$. As before, this may be optimized using first or second-order approaches. For intuition on the continued use of a single loss function $L$, consider Figure \ref{fig:RAPDARTS}. Under the assumption that a change in network parameters $\bm{\theta}$ creates no change in cost (given a fixed input feature map and primitive operation), the gradient of $C_1$ with respect to $\bm{\theta}$ is zero. On the other hand, a change in architecture parameters $\bm{\alpha}$ creates a change in both $L_1$ and $C_1$. So calculating the gradient of $L = L_1 + \lambda_1C_1$ with respect to both $\bm{\theta}$ and $\bm{\alpha}$ results in the correct values.

Using the method above, we created \textit{Resource-Aware P-DARTS} (RAPDARTS). Practically, the modification to P-DARTS requires the total expected resource cost be returned during the forward pass of an input tensor. To achieve this, during calculation of each mixed operation (Equation \ref{eq:expectedk}), we also calculate the expected resource cost (Equation \ref{eq:expectedr}). The expected cost for all mixed operations is accumulated and added to the classification loss (Equation \ref{eq:sumexp}). If multiple costs are required, e.g. model size and latency, each cost requires its own version of Equation \ref{eq:expectedr}, and must be accumulated individually from other costs.

\section{Experiments and Results}

We use RAPDARTS to search for CIFAR-10 neural architectures. We follow the architecture discovery algorithm of P-DARTS and search for cell architectures containing the same primitive operations as used by DARTS and P-DARTS, namely: 
    \begin{multicols}{2}
    \begin{itemize}
        \item Zero*
        \item Skip-Connect*
        \item Avg-Pool $3 \times 3$*
        \item Max-Pool $3 \times 3$*
        \item Seperable $3 \times 3$ Conv.
        \item Seperable $5 \times 5$ Conv.
        \item Dialated $3 \times 3$ Conv.
        \item Dialated $5 \times 5$ Conv.
    \end{itemize}
    \end{multicols}
All of the above primitive operations are standard convolutional layers except Zero which allows a cell to learn \textit{not} to pass information. Skip-connect is a parameter-free operation which allows information to pass through the mixed operation without modification. Parameter-free primitive operations are marked with an asterisk.

In an effort to simulate a real-world constraint, we restrict ourselves such that discovered CIFAR-10 architectures must have less than $3 \times 10^6$ parameters. This constrained optimization problem may be captured as:
\begin{equation}
\begin{aligned}
& \underset{\bm{\theta}, \bm{\alpha}}{\text{minimize}}
& & L_1(g(\bm{x}; \bm{\theta}, \bm{\alpha}), y) \\
& \text{subject to} & & \text{Parameter count} \leq 3 \times 10^6.
\end{aligned}
\label{eq:constraint}
\end{equation}
We perform NAS adhering to this constraint using the RAPDARTS framework above.

For the purpose of baseline calculations, we first consider the unconstrained results from P-DARTS. The authors of P-DARTS provided a reference architecture discovered through their algorithm \cite{pdartsgtype}. We trained and evaluated that architecture eight times using the latest version of the P-DARTS code \cite{pdartscode}. We then used the results from the repeated training to obtain performance statistics of the published architecture. 

The resulting trained models achieved $2.60 \pm .13\%$ error on the CIFAR-10 validation dataset. Additionally, the published P-DARTS architecture requires $3.4 \times 10^6$ parameters.

We then executed the P-DARTS architecture search code four times to test the ability to rediscover architectures with the performance of the published architecture. The four searches resulted in nine architectures. However, per the P-DARTS algorithm, we eliminated one architecture with more than two skip-connections in the \textit{normal} cell (see P-DARTS paper for details on the two cell types). 

None of the eight valid architectures were the same as the official P-DARTS CIFAR-10 architecture, but this is not surprising, given the size of the P-DARTS architecture search space. Because of this, we compare our results to the statistics of various architectures discovered during our search, instead of the statistics of the single published architecture. The resulting trained models achieved $2.72 \pm .22\%$ error on CIFAR-10. The architectures required $3.9 \pm .3$ M parameters. The smallest P-DARTS model required $3.4$ M parameters.

We now explore the impact of different hyperparameter values on the unconstrained multi-objective version of Equation \ref{eq:constraint}:
\begin{equation}
L = L_1 + \lambda_1 C_1,
\end{equation}
where $C_1$ is the sum of expected number of parameters in the model. As introduced in Equation \ref{eq:totalloss}, the $\lambda_1$ scalar is a hyperparameter which determines the relative importance of the resource cost explicitly and the relative importance of the accuracy of the network implicitly.

\begin{figure}[!t]
\centering
\includegraphics[width=3.45in]{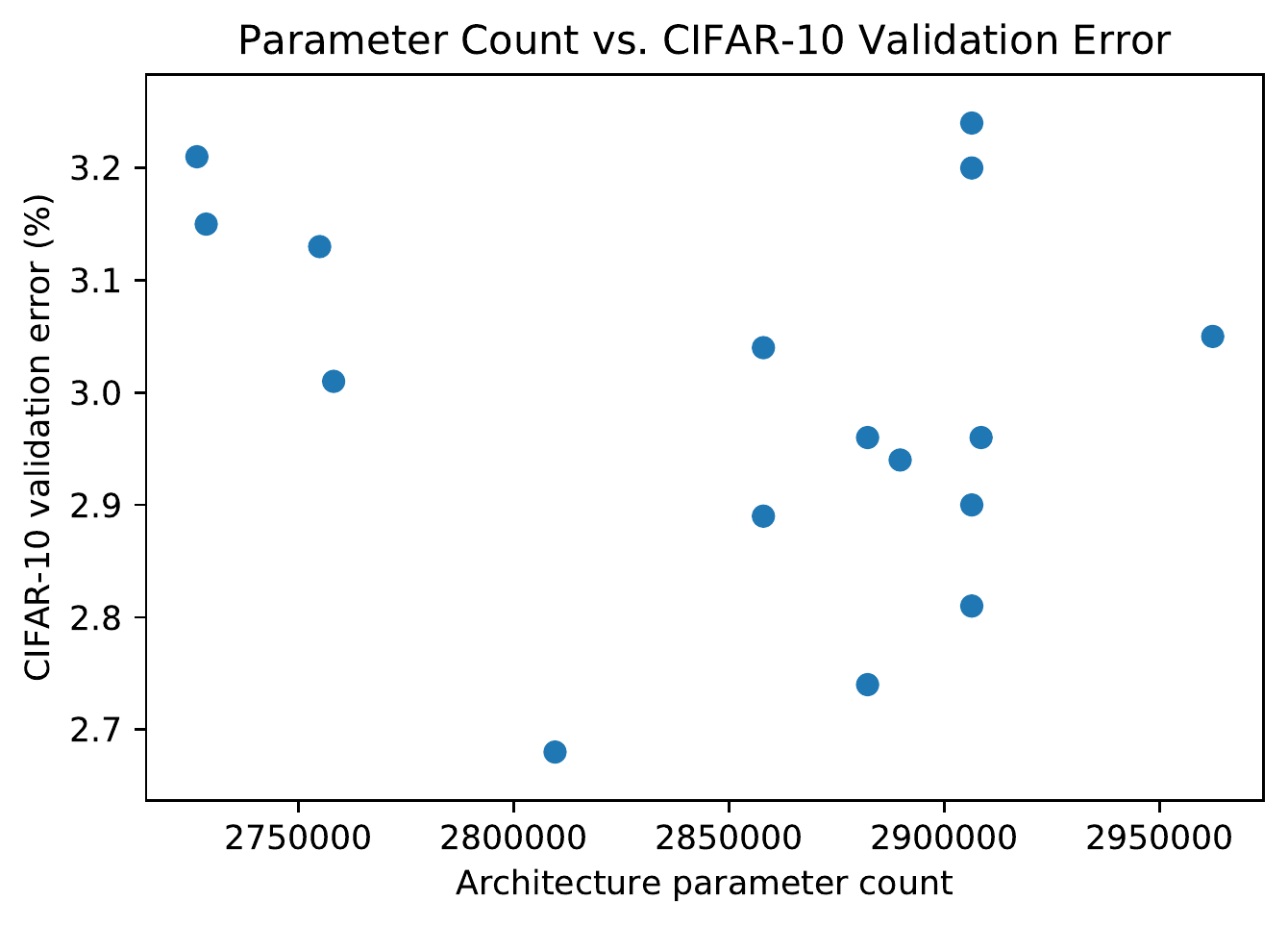}
\caption{Relationship between RAPDARTS model size and trained validation error appears uncorrelated. Indicating that at this variation of model capacity, model size is not a predictor of final classifier performance.}
\label{fig:fine}
\end{figure}

\begin{table*}[t!]
\centering
\begin{tabular}{llllll}
\hline
\textbf{}             & \multicolumn{2}{l}{\textbf{C10 Test Err (\%)}} &                     &                                 & \multicolumn{1}{c}{}   \\
\textbf{Architecture} & \textbf{Best}             & \textbf{Avg}             & \textbf{Params (M)} & \textbf{Search Cost (GPU-days)} & \textbf{Search Method} \\ \hline
AmoebaNet+B + cutout \cite{real2019regularized}          & N/A             & $\bm{2.55 \pm 0.05}$ & 2.8 & 3150 & evolution \\
ASHA  \cite{li2019random}                                & 2.85            & $3.03 \pm 0.13$          & 2.2 & 9    & random \\
DARTS \cite{liu2018darts}                                & 2.94            & N/A                      & 2.9 & .4   & gradient-based \\
DSO-NAS \cite{zhang2018you}                              & N/A             & $2.84 \pm 0.07$          & 3.0 & 1    & gradient-based \\
SNAS + moderate constraint + cutout \cite{xie2018snas}   & 2.85            & N/A                      & 2.3 & 1.5  & gradient-based \\ \hline
RAPDARTS + cutout (ours)                                 & 2.68            & $2.83 \pm 0.05$          & 2.8 &  12  & gradient-based \\ \hline \\
\end{tabular}
\caption{RAPDARTS CIFAR-10 error rate versus others for models with less than $3 \times 10^6$ parameters. We also include NAS results from randomly searched architectures \cite{li2019random}. For RAPDARTS, search cost includes actual cost for all experiments for finding the 2.68\% model. In total, the search and train phases required 26 GPU-days.}
\label{tab:my-table}
\end{table*}

As stated in this section's introduction, our self-imposed resource budget is 3 M parameters. The default P-DARTS search does not generate models that small, however, by using RAPDARTS we are able to satisfy this constraint. To achieve this, we need to discover a $\lambda_1$ value to guide the architecture search. That is accomplished by finding a coarse range of suitable $\lambda_1$s and then identifying a refined $\lambda_1$. 

The coarse $\lambda_1$ is identified by performing various architecture searches with $\lambda_1$s sampled randomly from a uniform distribution $\mathcal{U}([10^{-11},10^{-6}])$. Each search requires .3 GPU-days. 

\begin{figure}[ht]
\centering
\subfloat[Normal Cell]{%
  \includegraphics[width=3.45in]{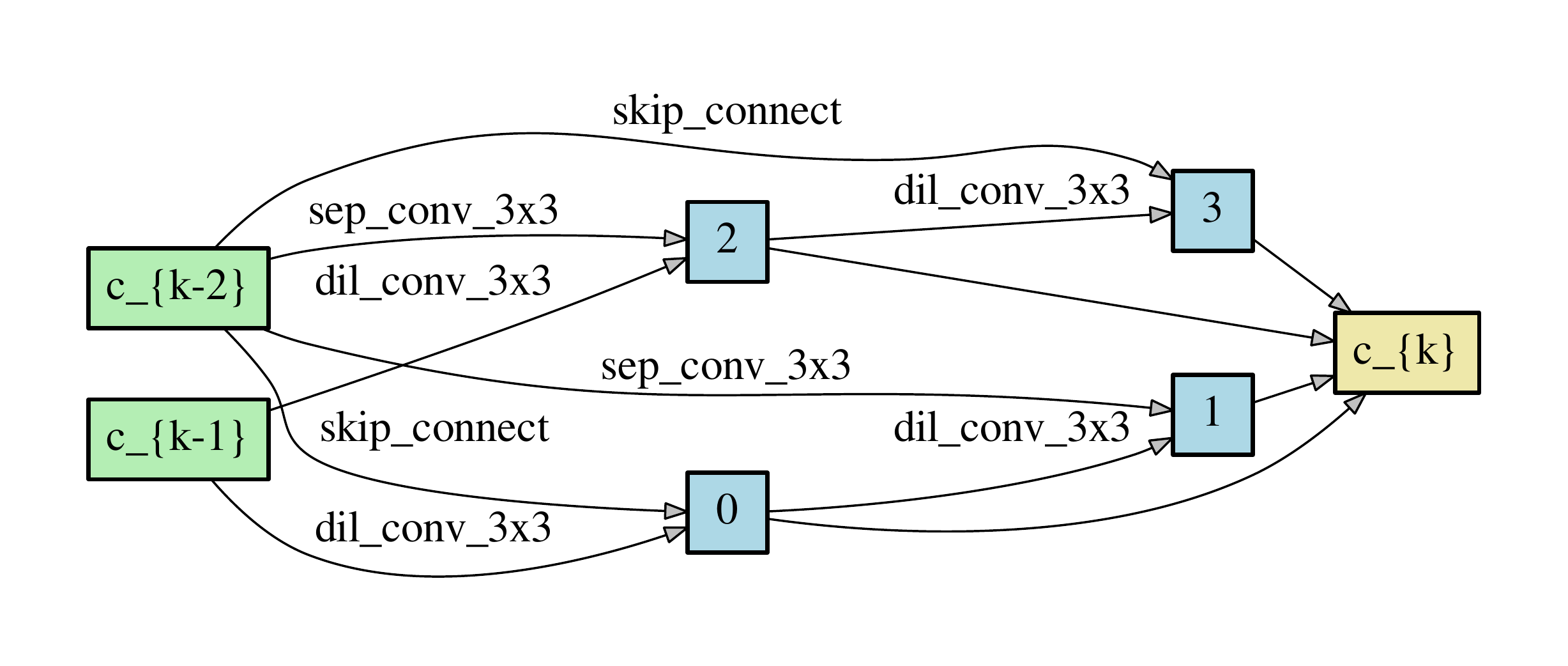}%
}

\subfloat[Reduce Cell]{%
  \includegraphics[width=2.35in]{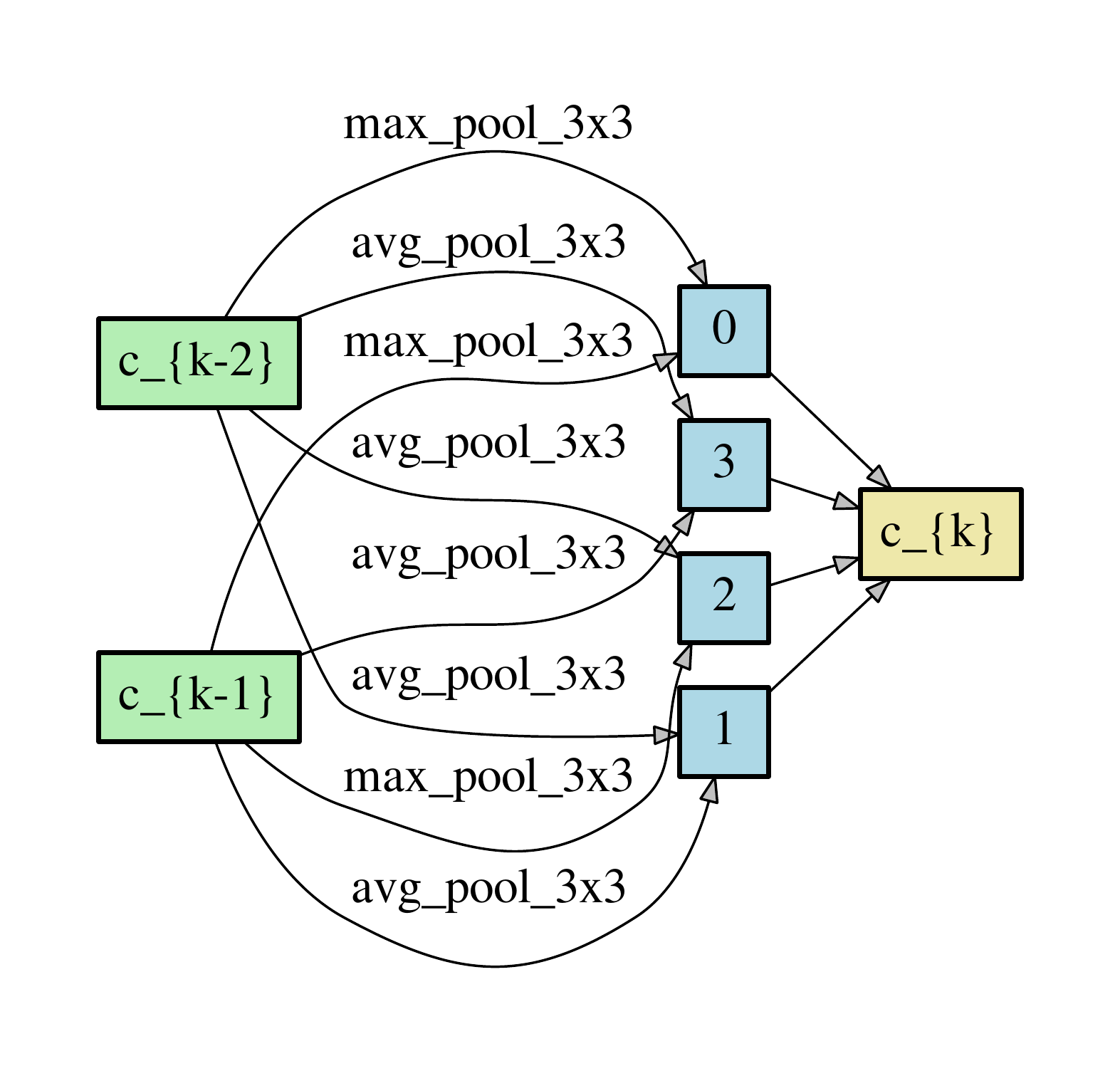}%
}

\caption{Cells found by RAPDARTS achieving 2.68\% CIFAR-10 validation error. All primitive operations are low-cost operations.}
\label{fig:discovered}
\end{figure}

Results from the coarse-search are shown in Figure \ref{fig:lam_bytes}. At approximately $\lambda_1 > 10^{-7}$, architectures begin to be generated which meet the $3 \times 10^{6}$ parameter count constraint. Parameter counts reduce dramatically as $\lambda_1$ approaches $10^{-6}$, but we have observed that models with higher capacity tend to perform better than models with lower capacity, so it is unlikely that architectures derived from $\lambda_1 > 10^{-6}$ are preferred over those closer to the 3 M parameter threshold.

Figure \ref{fig:lam_bytes_zoom} ``zooms in'' on the previous figure, focusing on $\lambda_1$ sampled uniformly from $\mathcal{U}[(2 \times 10^{-7}, 10^{-6})]$. Near $\lambda_1 = 6 \times 10^{-7} \approx 1 \times 10^{-6.2}$, architectures are generated that often require less than 3 M parameters. 

One final search is then performed on $\lambda_1$ sampled uniformly from $U([10^{-6.24},10^{-6.2}])$. This test resulted in 48 valid architectures with resulting models between 2.1 M and 2.96 M parameters. We then trained the 16 largest resulting architectures. The resulting best model achieved 2.68\% CIFAR-10 validation error and required 2.8 M parameters. The results for all 16 trained models are plotted in Figure \ref{fig:fine}. As can be seen, there is no linear relationship at this scale between parameter count and CIFAR-10 accuracy. For statistical confidence, we retrained the best model eight times with different seeds and obtained $2.83\% \pm .05$ validation error.

The discovered cells corresponding to the 2.68\% CIFAR-10 validation are shown in Figure \ref{fig:discovered}. The DARTS-based algorithms use two cell types: a ``normal'' cell, which maintains input and output feature map dimensionality, and a ``reduce'' cell, which decrease the output feature maps dimensionality. 

The cell architectures discovered by RAPDARTS are noteworthy in several respects. First, the normal cell has discovered a ``deep'' design, similar to that discovered by P-DARTS, but only light-weight convolutional operations are used. Second, all pooling operations have been moved to the reduce cells.

Table \ref{tab:my-table} compares the RAPDARTS architecture with the performance of recent architectures with parameter counts less than 3 M. RAPDARTS competes favorably with the others.

We report the actual number of hours spent searching for our winning architecture, not merely the search time for a single architecture. Including both the coarse and fine-search phases, 40 different $\lambda_1$ values were used. This took a total of 12 GPU-days to compute.

We trained 16 of the fine-search phase models to completion. Each model required less than 20 hours to train, so the 16 fine-search models took less than 14 GPU-days total to train. All experiments were performed using an NVIDIA V100 GPU.

\section{Conclusion and Future Work}

Classification accuracy achieved by neural architecture search methods now surpass hand-designed neural models. First-generation NAS methods include those based on evolutionary search and reinforcement learning. Second generation NAS methods use gradient-based optimization. In this work we present RAPDARTS, which augments a popular gradient-based NAS method with the ability to target neural architectures meeting specified resource constraints. We use RAPDARTS to identify a neural architecture achieving 2.68\% test error on CIFAR-10. This is competitive with other existing results for models with less than 3 M parameters.

We believe third-generation methods will be gradient-based and attempt to make more aspects of the search differentiable. For example, the P-DARTS (and RAPDARTS) search begins with five cells, then grows the search network to 11 cells, and finally 17 cells. At the same time, as the network grows, less important primitive operations are dropped. The ``gradual'' adjustments introduced by this technique enable architecture parameters learned by gradient-descent in one phase to be useful in another. It would be preferable to make these changes even more gradually. We leave that for future work.

In conclusion, we have presented an example that optimizes two objectives: minimizing accuracy loss while keeping the number of model parameters below a resource constraint threshold. A limitation of our work is that the number of parameters required by our discovered models may not optimize other constraints, e.g. minimum latency. To address this concern, future work will focus on multiple resource constraints guided by more hardware-specific costs.

\section*{Acknowledgment}
Sandia National Laboratories is a multi-mission laboratory managed and operated by National Technology and Engineering Solutions of Sandia, LLC., a wholly owned subsidiary of Honeywell International, Inc., for the U.S. Department of Energy's National Nuclear Security Administration under contract DE-NA0003525.

The views expressed in the article do not necessarily represent the views of the U.S. Department of Energy or the United States Government.

\ifCLASSOPTIONcaptionsoff
  \newpage
\fi

\bibliographystyle{IEEEtran}
\bibliography{references}

\end{document}